\documentclass{article} 
\usepackage{iclr2021_conference,times}

\usepackage{amsmath,amsfonts,bm}

\def\eqref#1{equation~\ref{#1}}

\def\1{\bm{1}}

\DeclareMathAlphabet{\mathsfit}{\encodingdefault}{\sfdefault}{m}{sl}
\SetMathAlphabet{\mathsfit}{bold}{\encodingdefault}{\sfdefault}{bx}{n}

\newcommand{\R}{\mathbb{R}}

\usepackage[para]{footmisc}
\usepackage{hyperref}
\usepackage[capitalise]{cleveref}
\usepackage{url}
\usepackage{graphicx}
\usepackage[inline]{enumitem}
\usepackage{bm}
\usepackage{amsfonts,amssymb,amsmath}
\usepackage{caption}
\usepackage{subcaption}
\usepackage{wrapfig}
\usepackage{pifont}

\newcommand{\cmark}{\ding{51}}
\newcommand{\xmark}{\ding{55}}

\title{CausalWorld: A Robotic Manipulation Benchmark for Causal Structure and Transfer Learning
}

\author{Ossama Ahmed \textsuperscript{1,*} 
Frederik Tr\"auble \textsuperscript{2,*} 
Anirudh Goyal \textsuperscript{3}
Alexander Neitz  \textsuperscript{2} \\
\textbf{
Yoshua Bengio \textsuperscript{3}
Bernhard Sch\"olkopf \textsuperscript{2}
Stefan Bauer \textsuperscript{2,$\dag$}
Manuel W\"uthrich \textsuperscript{2,$\dag$}
}}

\iclrfinalcopy 
\begin{document}

\let\footnote\relax\footnotetext{
\textsuperscript{*} Equal Contribution ,
\textsuperscript{$\dag$} Equal Advising,
\textsuperscript{1} ETH Zürich,
\textsuperscript{2} MPI Tübingen,
\textsuperscript{3} Mila, University of Montreal,\\
Corresponding author: \texttt{ossama.ahmed@mail.mcgill.ca, frederik.traeuble@tuebingen.mpg.de}
\textsuperscript{4} https://sites.google.com/view/causal-world/home
}

\maketitle

\begin{abstract}
Despite recent successes of reinforcement learning (RL), it remains a challenge for agents to transfer learned skills to related environments.
To facilitate research addressing this problem, we propose {\em CausalWorld}, a benchmark for causal structure and transfer learning in a robotic manipulation environment. The environment is a simulation of an open-source robotic platform, hence offering the possibility of sim-to-real transfer. 
Tasks consist of constructing 3D shapes from a given set of blocks - inspired by how children learn to build complex structures. 
The key 
strength of {\em CausalWorld} is that it provides a combinatorial family of such tasks with common causal structure and underlying factors (including, e.g., robot and object masses, colors, sizes).
The user (or the agent) may intervene on all causal variables, which 
allows for fine-grained control over how similar different 
tasks (or task distributions) are. One can thus easily define
training and evaluation distributions of a desired difficulty level, targeting
a specific form of generalization (e.g., only changes in appearance or object mass). 
Further, this common parametrization facilitates defining curricula by 
interpolating between an initial and a target task. While users may define their own task distributions, we present eight meaningful distributions as concrete benchmarks, ranging from simple to very challenging, all of which require long-horizon planning as well as precise low-level motor control. Finally, we provide baseline results for a subset of these tasks on distinct training curricula and corresponding evaluation protocols, verifying the feasibility of the tasks in this benchmark.\textsuperscript{4}

\end{abstract}

\section{Introduction}
\label{sec:introduction}
\begin{wrapfigure}{R}{0.35\textwidth}
\centering
\includegraphics[clip,
    trim=5.8cm 19cm 5.8cm 2.5cm, width=\linewidth]{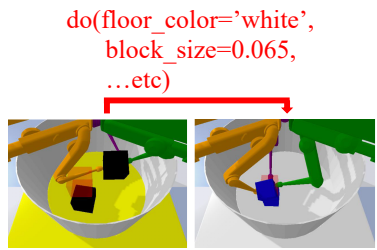}
\caption{\label{fig:stacking_task}Example of do-interventions on exposed variables in \texttt{CausalWorld}.}
\end{wrapfigure}

Benchmarks have played a crucial role in advancing entire research fields, for instance computer vision with the introduction of CIFAR-10 and ImageNet \citep{krizhevsky2009learning, krizhevsky2012imagenet}. When it comes to the field of reinforcement learning (RL), similar breakthroughs have been achieved in domains such as game playing \citep{mnih2013playing, silver2017mastering}, learning motor control for high-dimensional simulated robots \citep{akkaya2019solving},  multi-agent settings \citep{baker2019emergent, berner2019dota} and for studying transfer in the context of meta-learning \citep{yu2019meta}. Nevertheless, trained agents often fail to transfer the knowledge about the learned skills from a training environment to a different but related environment sharing part of the underlying task structure. This can be attributed to the fact that it is quite common to evaluate an agent on the training environments themselves, which leads to overfitting on these narrowly defined environments \citep{whiteson2011protecting}, or that algorithms are compared using highly engineered and biased reward functions which may result in learning suboptimal policies with respect to the desired behaviour;
this is particularly evident in robotics.

In existing benchmarks \citep{yu2019meta, goyal2019infobot, cobbe2018quantifying, bellemare2013arcade, james2020rlbench} the amount of shared causal structure between the different environments is mostly unknown. For instance, in the Atari Arcade Learning environments, it is unclear how to quantify the underlying similarities between different Atari games and we generally do not know to which degree an agent can be expected to generalize. To overcome these limitations, we introduce a novel benchmark in a robotic manipulation environment which we call \texttt{CausalWorld}. It features a diverse set of environments which, in contrast to previous designs, share a large set of parameters and parts of the causal structure. Being able to intervene on these parameters (individually or collectively) permits the experimenter to evaluate agents' generalization abilities with respect to different types and extents of changes in the environment. These parameters can be varied gradually, which yields a continuum of similar environments. This allows for fine-grained control of training and test distributions and the design of learning curricula.

\begin{figure}
    \captionsetup{position=bottom}
    \centering
    \begin{subfigure}{0.245\textwidth}
        \subcaption{Pushing}
        \includegraphics[width=\linewidth]{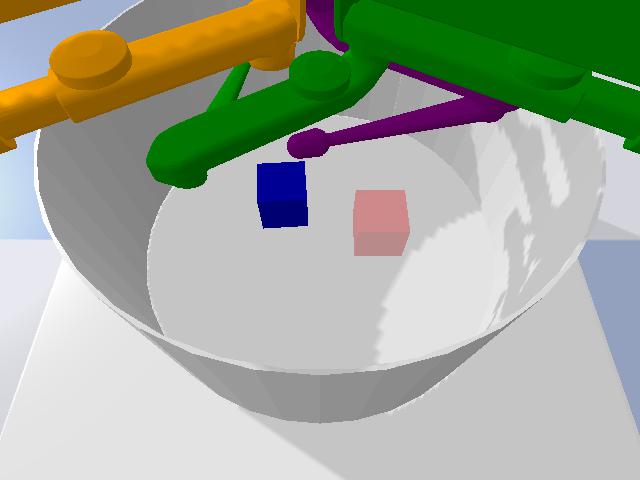}
    \end{subfigure}
        \begin{subfigure}{0.245\textwidth}
        \subcaption{Picking}
        \includegraphics[width=\linewidth]{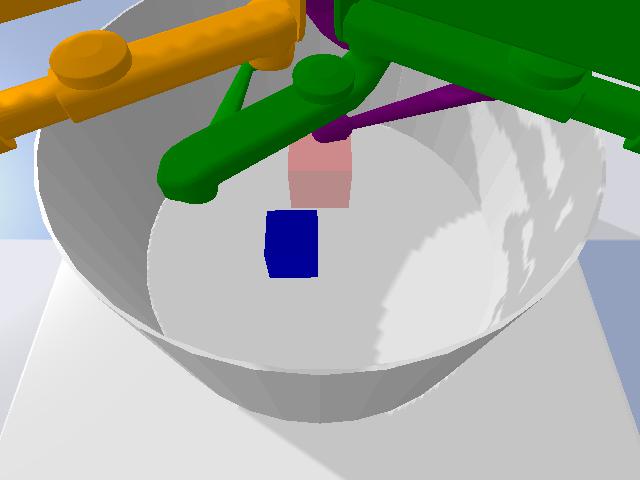}
    \end{subfigure}
        \begin{subfigure}{0.245\textwidth}
        \subcaption{Pick and Place}
        \includegraphics[width=\linewidth]{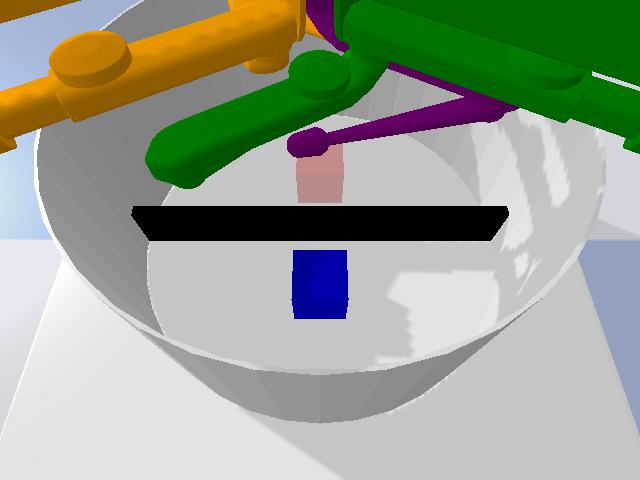}
    \end{subfigure}
    \begin{subfigure}{0.245\textwidth}
        \subcaption{Stacking2}
        \includegraphics[width=\linewidth]{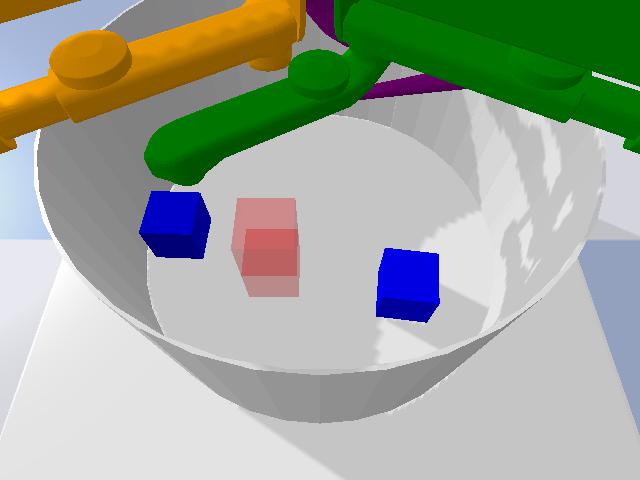}
    \end{subfigure}
    \begin{subfigure}{0.245\textwidth}
        \subcaption{Stacked Blocks}
        \includegraphics[width=\linewidth]{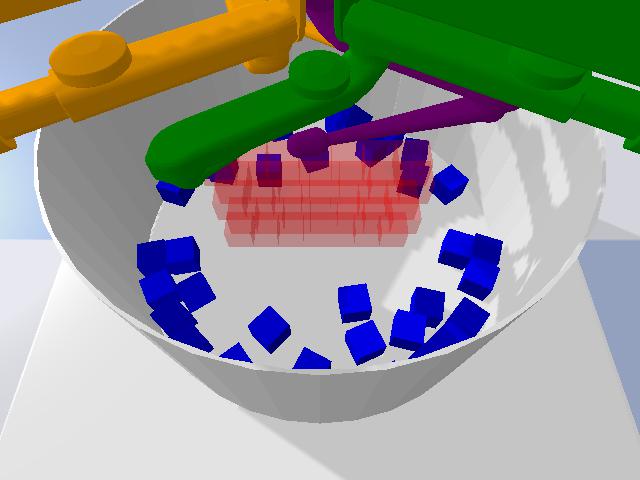}
    \end{subfigure}
        \begin{subfigure}{0.245\textwidth}
        \subcaption{General}
        \includegraphics[width=\linewidth]{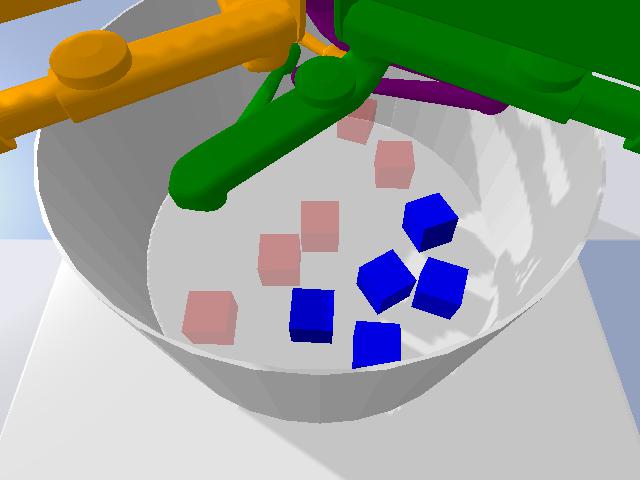}
    \end{subfigure}
        \begin{subfigure}{0.245\textwidth}
        \subcaption{CreativeStackedBlocks}
        \includegraphics[width=\linewidth]{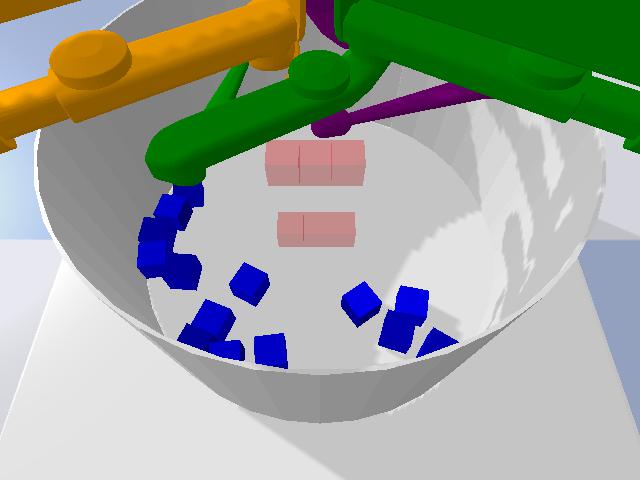}
    \end{subfigure}
    \begin{subfigure}{0.245\textwidth}
        \subcaption{Towers}
        \includegraphics[width=\linewidth]{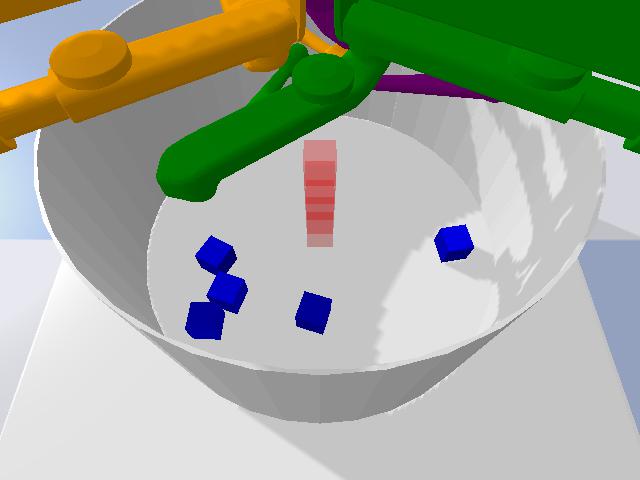}
    \end{subfigure}
    \qquad
    \caption{Example tasks from the task generators provided in the benchmark. The goal shape is visualized in opaque red and the blocks in blue.}
    \label{fig:task_snapshots}
\end{figure}

A remarkable skill that humans learn to master relatively early on in their life is building complex structures using their spatial-reasoning and dexterous manipulation abilities \citep{casey2008development, caldera1999children, kamii2004development}. Playing with toy blocks constitutes a natural environment for children to develop important visual-spatial skills, helping them `generalize' in building complex composition designs from presented or imagined goal structures \citep{verdine2017links, nath2014construction, dewar2018benefits, richardson2014children}. Inspired by this, \texttt{CausalWorld} is designed to aid in learning and investigating these skills in a corresponding simulated robotics manipulation environment of the open-source TriFinger robot platform from \cite{wuthrich2020trifinger}, which can be built in the real world. Tasks are formulated as building 3D goal shapes using a set of available blocks by manipulating them - as seen in \cref{fig:stacking_task}. 
This yields a diverse familiy of tasks, ranging from relatively simple (e.g. pushing a single object) to extremely hard (e.g. building a complex structure from a large number of objects).

\texttt{CausalWorld} improves upon previous benchmarks by exposing a large set of parameters in the causal generative model of the environments, such as weight, shape and appearance of the building blocks and the robot itself.  The possibility of intervening on any of these properties at any point in time allows one to set up training curricula or to evaluate an agent's generalization capability with respect to different parameters. Furthermore, in contrast to previous benchmarks \citep{chevalier2018babyai, cobbe2018quantifying}, researchers may build their own real-world platform of this simulator at low cost, as detailed in \cite{wuthrich2020trifinger}, and transfer their trained policies to the real world. 

Finally, by releasing this benchmark we hope to facilitate research in causal structure learning, i.e. learning the causal graph (or certain aspects of it) as we operate in a complex real-world environment whose dynamics follow the laws of physics which induce causal relations between the variables. Changes to the variables we expose can be considered do-interventions on the underlying structural causal model (SCM). Consequently, we believe that this benchmark offers an exciting opportunity to investigate causality and its connection to RL and robotics. 
 
Our main contributions can be summarized as follows:
\begin{itemize}
    \item  We propose \texttt{CausalWorld}, a new benchmark  comprising a parametrized family of robotic manipulation environments for advancing out-of-distribution generalization and causal structure learning in RL.
    \item We provide a systematic way of defining curricula and disentangling generalization abilities of trained agents by allowing do-interventions to be performed on all environment variables (parameters and states). 
    \item
    We establish baseline results for some of the available tasks under different learning algorithms, thus verifying the feasibility of the tasks.
    \item We show how different learning curricula affect generalization across different axes by reporting some of the in-distribution and out-of-distribution generalization capabilities of the trained agents.
\end{itemize}

\section{CausalWorld Benchmark}
\label{sec:benchmark}
\begin{table}[]
    \centering
    \tiny
    \begin{tabular}{p{6em}|p{4.5em}|p{5em}|p{5em}|p{3em}| p{4.5em}|p{4em}|p{3em}|p{3em}|p{3em}|p{3em}}
        Benchmark & do-interventions interface & procedurally generated environments & online distribution of tasks & setup custom curricula & disentangle generalization ability & real-world similarity &
        open-source robot & low-level motor control & long-term planning & unified success metric\\
        \hline\hline
        RLBench & \xmark & \xmark & \xmark & \xmark & \xmark & \cmark & \xmark &
        \cmark & \xmark & \xmark\\
        \hline
        MetaWorld & \xmark & \xmark & \xmark & \xmark & \xmark & \cmark &
        \xmark &
        \cmark & \xmark & \xmark\\
        \hline
        IKEA & \xmark & \xmark & \xmark & \xmark & \xmark & \cmark &
        \xmark &
        \cmark & \cmark & \cmark\\
        \hline
        MuJoBan & \xmark & \xmark & \xmark & \cmark & \xmark & \cmark &
        \xmark &
        \cmark & \cmark & \cmark\\
        \hline
        BabyAI & \xmark & \cmark & \xmark & \xmark &  \xmark & \xmark  &
        \xmark &
        \xmark & \cmark & \cmark\\
        \hline
        CoinRun & \xmark & \cmark & \xmark & \xmark & \xmark & \xmark &
        \xmark &
        \xmark & \xmark & \cmark\\
        \hline
        AtariArcade & \xmark & \xmark & \xmark & \xmark & \xmark & \xmark &
        \xmark &
        \xmark & \cmark/\xmark & \cmark\\
        \hline
        \texttt{CausalWorld} & \cmark & \cmark & \cmark & \cmark & \cmark & \cmark & \cmark & \cmark & \cmark & \cmark\\
    \end{tabular}
    \caption{Comparison of Causal World with RLBench \citep{james2020rlbench}, MetaWorld \citep{yu2019meta}, IKEA \citep{lee2019ikea}, BabyAI \citep{chevalier2018babyai}, CoinRun \citep{cobbe2018quantifying}, AtariArcade \citep{bellemare2013arcade}, MuJoBan etc. \citep{mirza2020physically}, }
    \label{tab:related_benchmark_table}
\end{table}

Here we make the desiderata outlined in the introduction more precise:

\begin{enumerate}
\item The set of environments should be sufficiently diverse to allow for the design of challenging transfer tasks.
\item We need to be able to intervene on different properties (e.g. masses, colors) individually, such that we can investigate different types of generalization.
\item It should be possible to convert any environment to any other environment by gradually changing its properties through interventions; this requirement is important for evaluating different levels of transfer and for defining curricula.
\item The environments should share some causal structure to allow algorithms to transfer the learned causal knowledge from one environment to another.
\item There should be a unified measure of success, such that an objective comparison can be made between different learning algorithms.
\item The benchmark should make it easy for users to define meaningful distributions of environments for training and evaluation. In particular, it should facilitate evaluation of in-distribution and out-of-distribution performance.
 \item The simulated benchmark should have a real-world counterpart to allow for sim2real.

\end{enumerate}

In light of these desiderata, we propose a setup in which a robot must build goal shapes using a set of available objects. It is worth noting that similar setups were proposed previously in a less realistic setting as in \citep{janner2018reasoning, bapst2019structured, mccarthylearning,akkaya2019solving, fahlman1974planning, winston1970learning, winograd1972understanding}. Specifically, a task is formulated as follows: \textit{given a set of available objects the agent needs to build a specific goal structure}, see \cref{fig:stacking_task} for an example. The vast amount of possible target shapes and environment properties (e.g. mass, shape and appearance of objects and the robot itself) makes this a diverse and challenging setting to evaluate different generalization aspects.  \texttt{CausalWorld} is a simulated version (using the Bullet physics engine \citep{coumans2013bullet}) of the open-source TriFinger robot platform from \citet{wuthrich2020trifinger}. Each environment is defined by a set of variables such as gravity, floor friction, stage color, floor color, joint positions, various block parameters (e.g. size, color, mass, position, orientation), link colors, link masses and the goal shape. See \cref{tab:causal_world_variables} in the Appendix for a subset of these variables.

Desideratum 1 is satisfied since different environment properties and goal shapes give rise to very different tasks, ranging from relatively easy (e.g. re-positioning a single cube) to extremely hard (e.g. building a complex structure). Desideratum 2 is satisfied because we allow for arbitrary interventions on these properties, hence users or agents may change parameters individually or jointly. Desideratum 3 is satisfied because the parameters can be changed gradually. Desideratum 4 is satisfied because all the environments share the causal structure of the robot, and one may also use subsets of environments which share even more causal structure. We satisfy desideratum 5 by defining the measure of success for all environments as the volumetric overlap of the goal shape with available objects. Further, by splitting the set of parameters into a set A, intended for training and in-distribution evaluation, and a set B, intended for out-of-distribution evaluation, we satisfy desideratum 6. Finally, since the TriFinger robot \citep{wuthrich2020trifinger} can be built in the real-world, we satisfy desideratum 7. Desideratum 7 and 2 are in partial conflict since sim2real is only possible for the tasks which are constrained to the variables on which the robot can physically act upon.


\paragraph{Task generators:} \label{task_generators} 
To generate meaningful families of similar goal shapes, \texttt{CausalWorld} allows for defining task generators which can generate a variety of different goal shapes in an environment. For instance, one task generator may generate pushing tasks, while another one may generate tower-building tasks (see \cref{fig:task_snapshots}).
Each task generator is initialized with a default goal shape from its corresponding family and comes with a sampler to sample new goal shapes from the same family. Additionally, upon construction, one can specify the environments' initial state and initial goal shape structure when deviating from the default. The maximum episode time to build a given shape is $\texttt{number\_of\_blocks} \times 10$ seconds.
\texttt{CausalWorld} comes with eight pre-defined task generators (see \cref{fig:task_snapshots}). 


\begin{itemize}
    \item Three generators create goal shapes with a single block: \textit{Pushing} with the goal shape on the floor, \textit{Picking} having the goal shape defined above the floor and \textit{Pick and Place} where a fixed obstacle is placed between the initial block and goal pose.
    \item \textit{Stacking2} involves a goal shape of two stacked blocks, which can also be considered one instance of the \textit{towers} generator.
    \item The remaining generators use a variable number of blocks to generate much more complex and challenging target shapes with details in the appendix: \textit{Towers}, \textit{Stacked Blocks}, \textit{Creative Stacked Blocks} and \textit{General}.
\end{itemize}

Given that building new environments using current physics simulators is often tedious, we provide a simple API for users who wish to create new task generators, for new challenging shape families which may be added to \texttt{CausalWorld's} task generators repository.

\paragraph{Action and Observation Spaces:}
The robot's action space $A \in \R^9$ can be chosen to operate in either joint position control mode, joint torque control mode, end-effector position control mode, or the delta of each. To address the different challenges in using high-dimensional visual observations as well as using a structured representation, we provide two observation modes: \textit{structured}  as well as \textit{pixel}. In the \textit{structured} mode, the low-dimensional observation vector $o$ follows a common rule for the ordering of the relevant variables, such as joints position, joints velocity, blocks linear velocity, time left for the task..etc. Thus, the observation space size depends on the number of blocks, which could potentially change with every new goal sampled, e.g. in Towers, (Creative) Stacked Blocks and General; therefore $O \in \R^{d_o}$ where $d_o$ varies across different environments. On the contrary, in the \textit{pixel} mode, the agent receives six different RGB images, where $O \in \R^{6\times 3\times 128\times 128}$, the first three images are rendered from different cameras mounted around the TriFinger robot, and the last three images specify the goal image of the target shape rendered from the same cameras. This mode can be mirrored on the real robotic platform and aids in investigating object-based learning approaches from pixel data as well as learning visual goal conditioned policies. Additionally, \texttt{CausalWorld} allows for setting up a fully customized observation space, if needed.

\paragraph{Rewards:} The reward function $r$ is defined uniformly across all possible goal shapes as the fractional volumetric overlap of the blocks with the goal shape, which ranges between 0 (no overlap) and 1 (complete overlap). This shared success metric can be returned at each time step where its scale is independent of the goal shape itself. Thus, an agent that learned this shared reward function $r$ from several different tasks could in principle use it to solve unseen goal structures. There is also the possibility of modifying the reward function to 1) sparsify the reward further by returning a binary reward signal instead, or 2) add a dense reward function in order to introduce inductive biases via domain knowledge and solution guidance. We hope that the considerable complexity and diversity of goal shapes motivate and accelerate the development of algorithms that are not dependent on highly tuned reward functions anymore.

\begin{figure}[!ht]
       \centering \includegraphics[clip,
        trim=0cm 4.5cm 0cm 11.3cm, width=\linewidth]{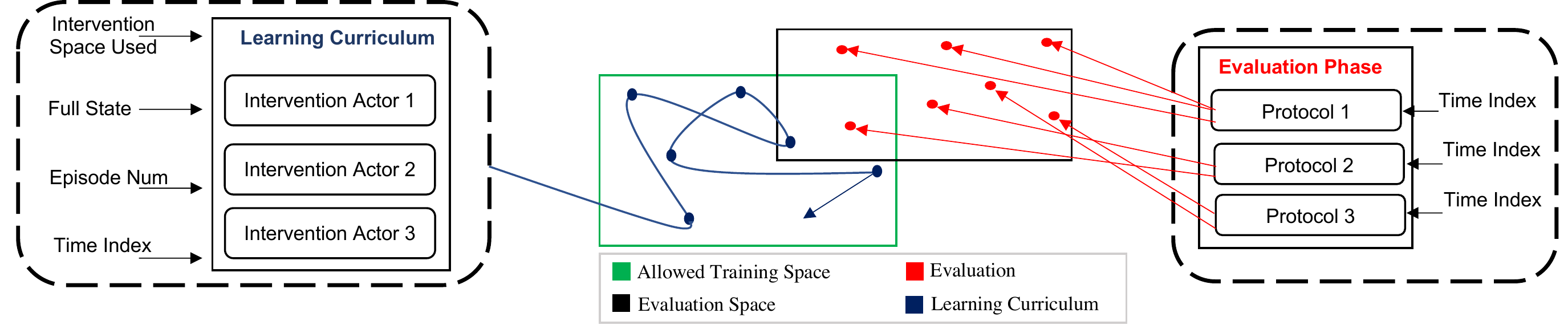}
\caption{\label{fig:interventions}
Key components for generic training and evaluation of RL agents. Left: A learning curriculum which is composed of various intervention actors that decide on which variables to intervene on (for a valid intervention, values need to be in the allowed training space (ATS)). Right: Evaluation protocols are shown which may intervene on variables at episode resets or within episodes (for a valid intervention, values need to be in the evaluation space (ES)). Middle: we represent the ATS and ES, where each intervention results in one point in the spaces. As shown ATS and ES may intersect, eg. if the protocols are meant to evaluate in-distribution generalization. A learning curriculum is represented by subsequent interventions navigating the ATS resulting in the corresponding points in the space.}

\end{figure}

{

\paragraph{Training and evaluation spaces:}
In this benchmark, a learning setting consists of an allowed training space (ATS) and an evaluation space (ES), both of which are subspaces of the full parameter space.
During training, in the simplest setting, parameters are sampled iid from the ATS. However, unlike existing benchmarks, \texttt{CausalWorld} allows for curricula within the ATS as well as settings where the agent itself intervenes on the parameters within an episode (see \cref{fig:interventions}). Similarly, during evaluation, parameters may be sampled iid from the evaluation space at each episode reset, or there can be interventions within an episode. Moreover, in order to retrieve the setting considered in most RL benchmarks, we could set the ATS and the ES to be identical and intervene only on object and robot states (and keep other environment properties constant) at each episode reset. However, to evaluate out-of-distribution generalization, one should set the two spaces (ATS and ES) to be different; possibly even disjoint. Additionally, to evaluate robustness with respect to a specific parameter (e.g. object mass), one may define the training and evaluation spaces to only differ in that particular parameter. In order to facilitate the definition of appropriate training and evaluation settings, we pre-define two disjoint sets, $\mathbf{A}_i$ and $\mathbf{B}_i$, for each parameter $i$. Through this, one can for instance define the training space to be $\mathbf{A}_1\times \mathbf{A}_2\times ...$ and the evaluation space to be $\mathbf{B}_1\times \mathbf{B}_2\times ...$ to assess generalization with respect to all parameters simultaneously. Alternatively, the evaluation space could be defined as $\mathbf{A}_1\times \mathbf{A}_2\times ... \times  \mathbf{B}_i \times \mathbf{A}_{i+1}\times ...$ to assess generalization with respect to parameter $i$ only. Lastly, users may also define their own spaces which could then be integrated into the benchmark to give rise to new learning settings.

}

\paragraph{Intervention actors:}
To provide a convenient way of specifying learning curricula, we introduce intervention actors. At each time step, such an actor takes all the exposed variables of the environment as inputs and may intervene on them. To encourage modularity, one may combine multiple actors in a learning curriculum. This actor is defined by the episode number to start intervening, the episode number to stop intervening, the timestep within the episode it should intervene and the episode periodicity of interventions. We provide a set of predefined intervention actors, including an actor which samples parameters randomly at each episode reset, which corresponds to domain-randomization. It is also easy to define custom intervention actors, we hope that this facilitates investigation into optimal learning curricula (see \cref{fig:interventions}).

\section{Related Work}
\label{sec:related_work}
Previous benchmarks proposed for RL mostly focused on the single task learning setting such as OpenAI Gym and DM control suite \citep{tassa2018deepmind, brockman2016openai}. Although, a recent line of work, e.g.\ Meta-World and RLBench \citep{yu2019meta, james2020rlbench} aim at studying multi-task learning as well as meta-learning, respective benchmarks mostly exhibit non-parametric hand-designed task variations which makes it ambiguous and not explicit how much structure is shared between them. For instance, it is not clear how different it is to ``open a door" compared to ``opening a drawer". To address the ambiguity in the shared structure between the tasks, \texttt{CausalWorld} was designed to allow interventions to be performed on many environment variables giving rise to a large space of tasks with well-defined relations between them, which we believe is a missing key component to address generalization in RL.

Similar parametric formulations of different environments were used in experiments in the generalization for RL literature, which have played an important role in advancing the field \citep{packer2018assessing, rajeswaran2017towards, pinto2017robust, yu2017preparing, henderson2017benchmark, dulac2020empirical, chevalier2018babyai}. In these previous works, variables were mostly assigned randomly as opposed to the full control over the variables in \texttt{CausalWorld} by allowing do-interventions.

Another important remaining challenge for the RL community is the standardization of the reported learning curves and results. RL methods have been shown to be sensitive to a range of different factors \citep{henderson2017deep}. Thus it is crucial to devise a set of metrics that measure reliability of RL algorithms and ensure their reproducibility. \citet{chan2019measuring} distinguishes between several evaluation modes like "evaluation during training" and "evaluation after learning". \citet{osband2019behaviour} recently proposed a benchmarking suite that disentangles the ability of an algorithm to deal with different types of challenges. Its main components are: enforcing a specific methodology for an agent's evaluation beyond the environment definition and isolating core capabilities with targeted 'unit tests' rather than integrating the general learning ability.

Moreover, causality has been historically studied from the perspective of probabilistic and causal reasoning \citep{pearl2009causality}, cognitive psychology \citep{griffiths2005structure}, and more recently in the context of machine learning
\citep{goyal2019recurrent, scholkopf2019causality, baradel2019cophy, bakhtin2019phyre}. On the contrary, we believe its link to robotics is not yet drawn systematically. To bridge this gap, one of the main motivations of \texttt{CausalWorld} was to facilitate research in causal learning for robotics, such as the capacity for observational discovery of causal effects in physical reality, counterfactual reasoning and causal structure learning.

\section{Experiments}
\label{sec:experiment}
To illustrate the usage of this benchmark and to verify the feasibility of some basic tasks, we evaluate current state-of-the-art model-free (MF-RL) algorithms on a subset of the goal shape families described in \cref{task_generators} and depticed in \cref{fig:task_snapshots}: (a) Pushing, (b) Picking, (c) Pick and Place, and (d) Stacking2. These goal shapes reflect basic skills that are required to solve more complex construction tasks.

\paragraph{Setup:} 
The idea here is to investigate how well an agent will perform on different evaluation distributions, depending on the curriculum it has been trained with. 
We train each method under the following curricula:

\begin{itemize}
    \item Curriculum 0: no environment changes; each episode is initialized from the default task lying in space A - note that here the initial state never changes (i.e. no interventions).
    \item Curriculum 1: goal shape randomization; at the beginning of each episode a new goal shape is sampled from space $\mathbf{A}$ (i.e. interventions on goal position and orientation).
    \item Curriculum 2: full randomization w.r.t. the task variables\textsuperscript{5}; every episode a simultaneous intervention on all variables is sampled from space $\mathbf{A}$ (i.e. can be seen as equivalent to extreme domain randomization in one space).
\end{itemize}
\let\footnote\relax\footnotetext{
\textsuperscript{5} Note that each task generator can suppress interventions that would yield goal shapes outside its family.
}
The curriculum will, as expected, affect the generalization capabilities of the trained agents. With \texttt{CausalWorld}'s formulation, these generalization capabilities can easily be disentangled and benchmarked quantitatively, as explained in \cref{sec:benchmark}. For each of the goal shape families (a, b, c, d from \cref{fig:task_snapshots}), we train agents under the three described curricula using the following MF-RL algorithms: The original Proximal Policy Optimization (PPO) from \citet{schulman2017proximal}, Soft Actor-Critic (SAC) from \citet{haarnoja2018soft} and the Twin Delayed DDPG (TD3) from \citet{fujimoto2018addressing}. We provided these methods with a hand-designed dense reward function as we did not observe any success with the sparse reward only.  Each of the mentioned setups is trained for five different random seeds, resulting in 180 trained agents. 
\begin{figure}
    \centering
    \includegraphics[width=\linewidth]{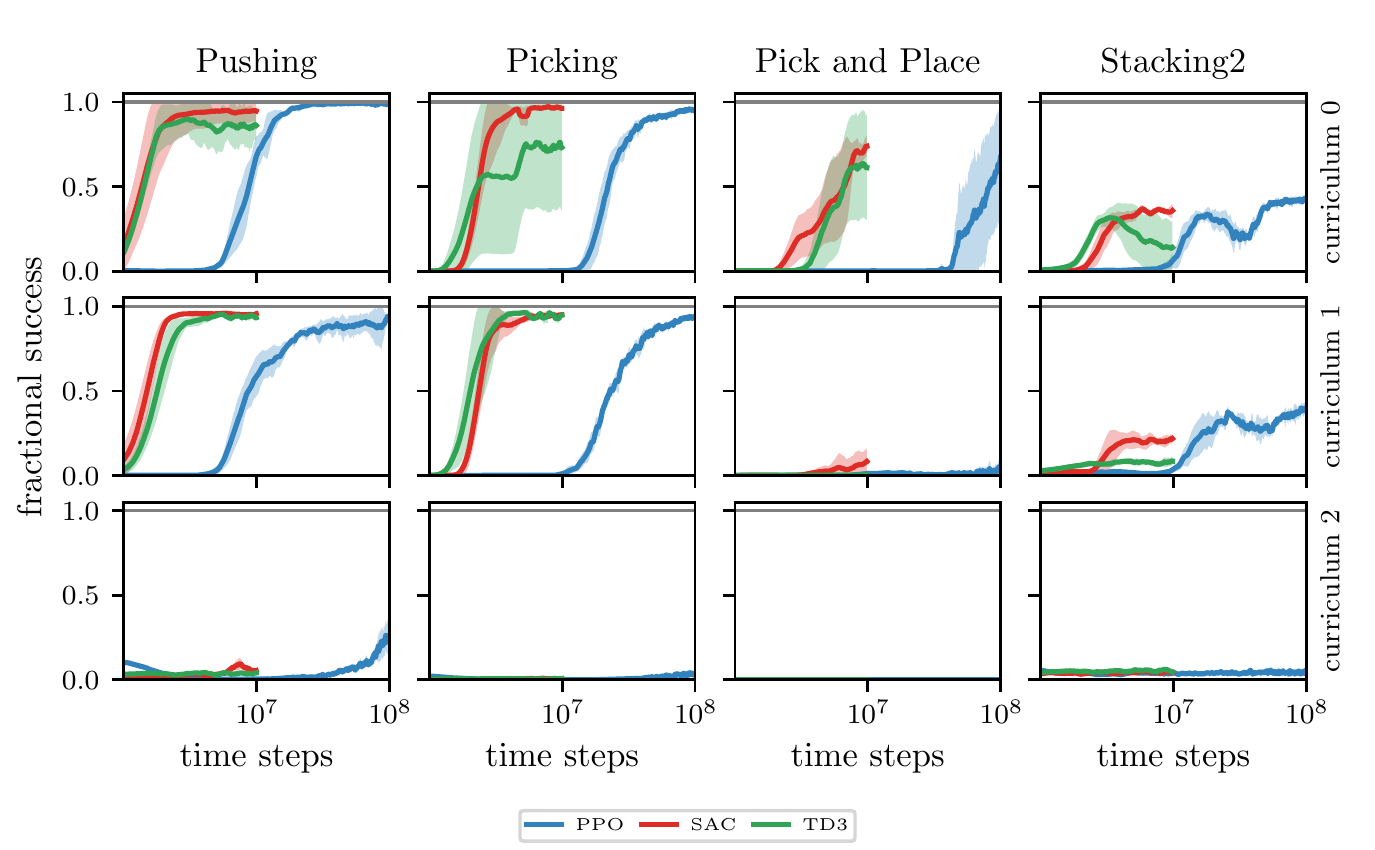}
    \caption{Fractional success curves averaged over five random seeds for the  tasks and learning algorithms specified above, under three different training curricula: (0) no curriculum, (1) goal position and orientation randomization in space $\mathbf{A}$ every episode and (2) a curriculum where we intervene on all variables in space $\mathbf{A}$ simultaneously every episode.}
    \label{fig:reward_curves}
\end{figure}
\paragraph{Training model-free RL methods:} We report the training curves averaged over the random seeds in \cref{fig:reward_curves}. As can be seen from these fractional success training curves, MF-RL methods are capable of solving the single block goal shapes (pushing, picking, pick and place) seen during training time given enough experience. However, we observe that none of the methods studied here managed to solve stacking two blocks. The score below 0.5 indicates that it only learns to push the lower cube into the goal shape. This shows that multi-object target shapes can become nontrivial quickly and that there is a need for better methods making use of the modular structure of object-based environments. To no surprise, the training curriculum has a major effect on learning, but the interpretation of generalization capabilities becomes much more explicit in the following subsection. For example, methods rarely manage to pick up any significant success signal under full extreme domain randomization as in curriculum 2, even after 100 million timesteps. Note that these curves represent the scores under the shapes and conditions of the actual training environments. Therefore, we need the capability of setting different protocols, in other words standardized sets of evaluation environment, that allow to benchmark learned skills of different agents.

\begin{figure}[!h]
    \centering
    \includegraphics[width=\linewidth]{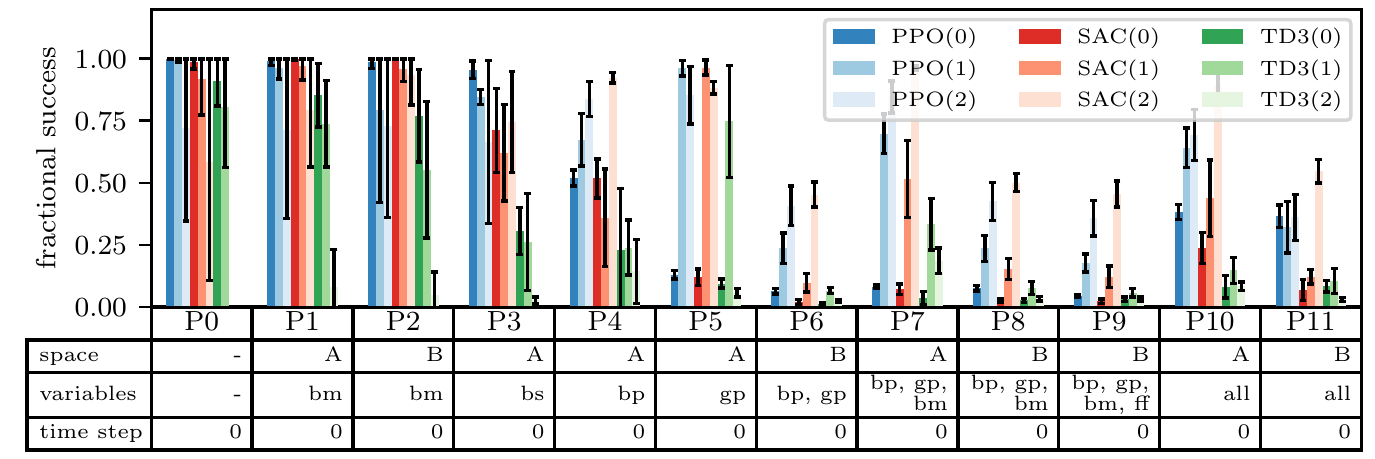}
    \caption{Evaluation scores for pushing baselines. Each protocol was evaluated for 200 episodes and each bar is averaged over five models with different random seeds. The variables listed under each protocol are sampled from the specified space at the start of every episode while all other variables remain fixed [bp block pose, bm block mass, bs block size, gp goal pose, ff floor friction].}
    \label{fig:evaluation_protocols_picking}
\end{figure}

\paragraph{Benchmarking generalization capabilities along various axes:} For each of the four goal shape families, we define a set of 12 evaluation protocols that we consider meaningful and representative for benchmarking the different algorithms. In the protocols presented here, we sample the values from a protocol-specific set of variables at the start of each episode while keeping all other variables fixed to their default values. After evaluating an agent on 200 episodes, we compute the fractional success score at the last time step of each episode and report the mean. These evaluation protocols allow to disentangle generalization abilities, as they show robustness with respect to different types of interventions, see \cref{fig:evaluation_protocols_picking}. The following are some of the observations we made for pushing:

\begin{itemize}
    \item Agents that were trained on the default pushing task environment (curriculum 0) do well (as expected) on the default task (P0). Interestingly, we likewise see a generalization capability to initial poses from variable space A (P4). This can be explained by a substantial exploration of the block positions via manipulation during training. Similarly, we see that the agents exhibit weaknesses regarding goal poses (P5) but overfit on their training settings instead.
    
    \item For agents trained with goal pose randomization (curriculum 1) we see similar results as with curriculum 0,  with the difference that agents under this curriculum generalize robustly to different goal poses (P5), as one would expect.
    \item Finally, agents that experience extreme domain randomization (curriculum 2) at training time, fail to learn any relevant skill as shown by the flat training curve in \cref{fig:reward_curves}. An explanation for this behavior could be that the agent might need more data and optimization steps to handle this much more challenging setting. Another possibility is that it may simply not be possible to find a strategy which simultaneously works for all parameters (note that the agent does not have access to the randomized parameters and hence must be robust to them). This poses an interesting question for future work.

\end{itemize}  

As expected, we observe that an agent's generalization capabilities are related to the experience gathered under its training curriculum. \texttt{CausalWorld} allows us to explore this relationship in a differentiated manner, assessing which curricula lead to which generalization abilities. This will not only help uncover an agent's shortcomings but may likewise aid in investigating novel learning curricula and approaches for robustness in RL.
Lastly, we note that this benchmark comprises extremely challenging tasks that appear to be out of reach of current model free methods without any additional inductive bias.

\section{Conclusion}
\label{sec:conclusion}
We have introduced a new benchmark - \texttt{CausalWorld} - to accelerate research in causal structure and transfer learning using a simulated environment of an open-source robot, where learned skills could potentially be transferred to the real world. We showed how allowing for interventions on the environment's properties yields a diverse familiy of tasks with a natural way of defining learning curricula and evaluation protocols that can disentangle different generalization capabilities.

We hope that the flexibility and modularity of \texttt{CausalWorld} will allow researchers to easily define appropriate benchmarks of increasing difficulty as the field progresses, thereby coordinating research efforts towards ever new goals.

\section{Acknowledgments}
\label{sec:acknowledgments}

The authors would like to thank Felix Widmaier, Vaibhav Agrawal and Shruti Joshi for the useful discussions and for the development of the TriFinger robot's simulator \citep{trifinger-simulation}, which served as a starting point for the work presented in this paper. AG is also grateful to Alex Lamb and Rosemary Nan Ke for useful discussions. The authors are grateful for the support from CIFAR. 


\bibliography{iclr2021_conference}
\bibliographystyle{iclr2021_conference}

\clearpage

\section{Appendix}
\appendix
\section{Observations}

Observations in \texttt{CausalWorld} has two modes, "structured" and "pixel". When using "pixel" mode, 6 images are returned consisting of the current images rendered from 3 different views on top of the TriFinger platform, showing the current state of the environment, as well as the 3 equivalent goal images rendered from the same points of view, showing the goal shape that the robot have to build by the end of the episode.
\begin{figure}[!ht]
    \captionsetup{position=bottom}
    \centering
    \begin{subfigure}{0.33\textwidth}
        \subcaption{Current View60}
        \includegraphics[width=\linewidth]{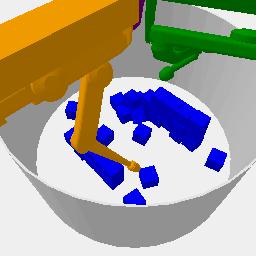}
    \end{subfigure}
        \begin{subfigure}{0.33\textwidth}
        \subcaption{Goal View60}
        \includegraphics[width=\linewidth]{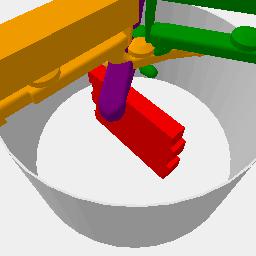}
    \end{subfigure}
        \begin{subfigure}{0.33\textwidth}
        \subcaption{Current View120}
        \includegraphics[width=\linewidth]{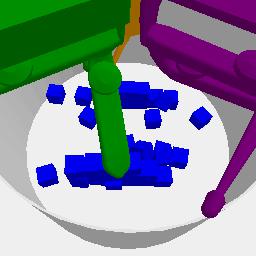}
    \end{subfigure}
    \begin{subfigure}{0.33\textwidth}
        \subcaption{Goal View120}
        \includegraphics[width=\linewidth]{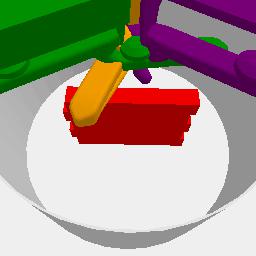}
    \end{subfigure}
    \begin{subfigure}{0.33\textwidth}
        \subcaption{Current View300}
        \includegraphics[width=\linewidth]{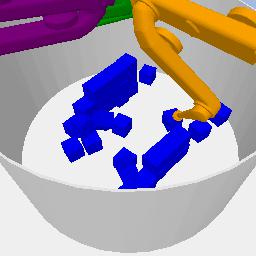}
    \end{subfigure}
        \begin{subfigure}{0.33\textwidth}
        \subcaption{Goal View300}
        \includegraphics[width=\linewidth]{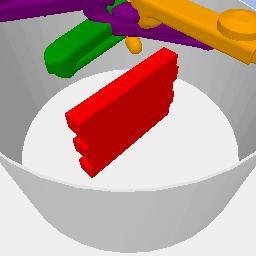}
    \end{subfigure}

    \qquad
    \caption{Example "pixel" mode observations returned at each step of the environment.}
    \label{fig:pixel_observations}
\end{figure}

\begin{figure}[!ht]
\centering
 \includegraphics[clip,
    trim=3cm 2.8cm 8.8cm 10cm, width=\linewidth]{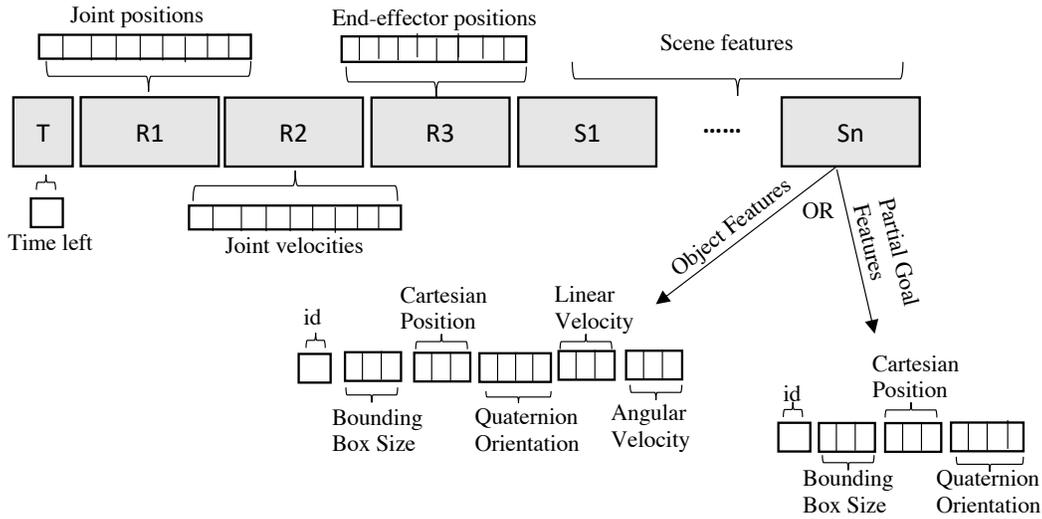}
    \caption{Structured observation description. For the scene features, all the blocks feature vector are concatenated first. Following that the partial goals feature vector are concatenated in the same order. Lastly, if there is any obstacles/ fixed blocks, their feature vectors are concatenated at the end following the same description as the partial goal features.}
    \label{fig:trifinger}
\end{figure}
\section{TriFinger Platform}
The robot from \citep{wuthrich2020trifinger} shown in figure \ref{fig:trifinger} is open-sourced and can be reproduced and built in any research lab; since its inexpensive (about \$5000), speeding up sim2real research.

\begin{figure}[!ht]
\centering
 \includegraphics[width=0.5\linewidth]{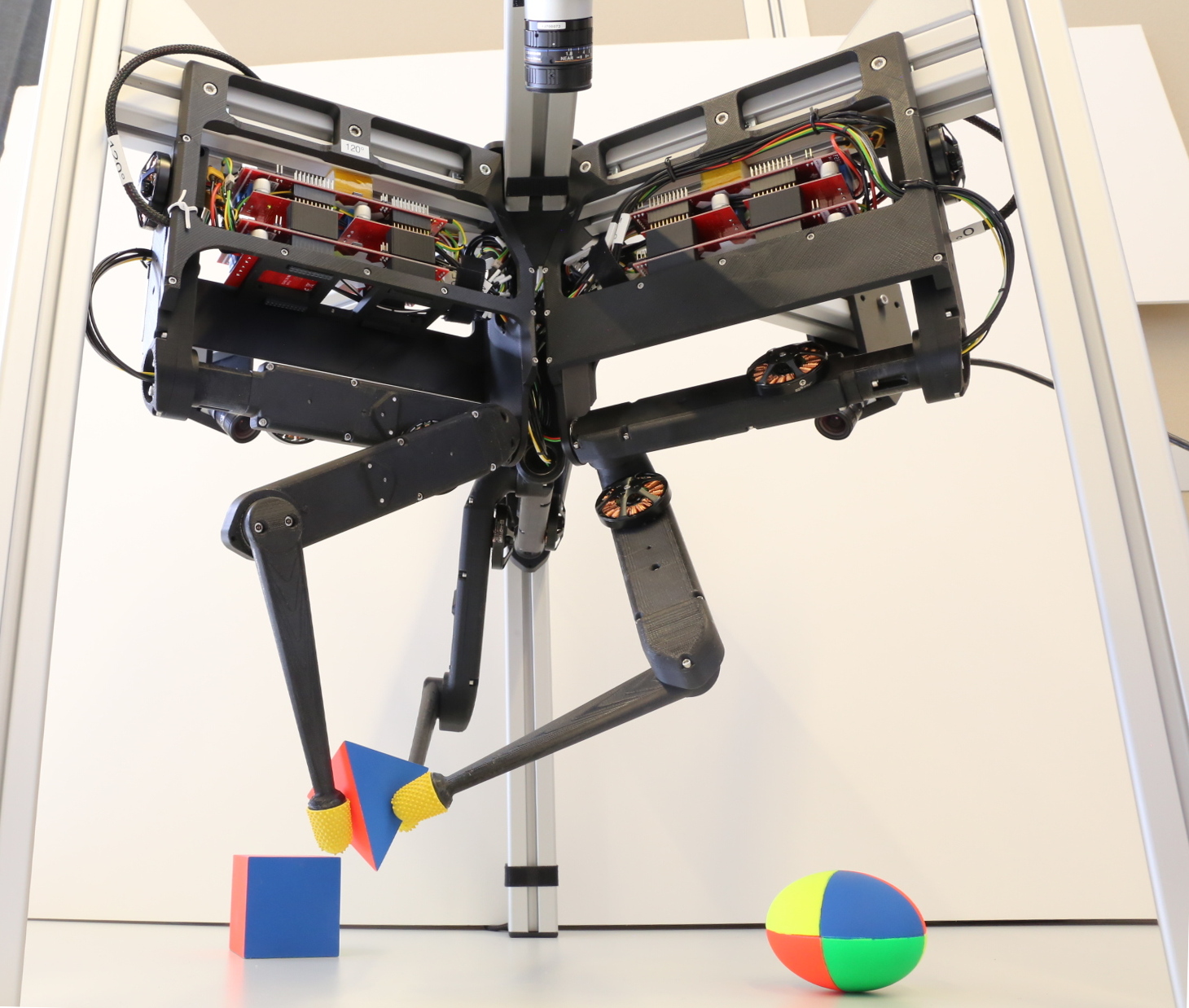}
    \caption{The TriFinger platform.}
    \label{fig:trifinger}
\end{figure}
\section{Task Generators}

\begin{enumerate}
    \item \textit{Pushing}: task where the goal is to push one block towards a goal position with a specific orientation; restricted to goals on the floor level.
    \item \textit{Picking}: task where the goal is to pick one block towards a goal height above the center of the arena; restricted to goals above the floor level.
    \item \textit{Pick And Place}: task where the arena is divided by a fixed long block and the goal is to pick one block from one side of the arena to a goal position with a variable orientation on the other side of the fixed block.
    \item \textit{Stacking2}: task where the goal is to stack two blocks above each other in a specific goal position and orientation.
    \item \textit{Towers}: task where the goal is to stack multiple n blocks above each other in a specific goal position and orientation - exactly above each other creating a tower of blocks.
    \item \textit{Stacked Blocks}: task where the goal is to stack multiple n blocks above each other in an arbitrary way to create a stable structure. The blocks don't have to be exactly above each other; making it more challenging than the ordinary towers task since the its harder to come up with a stable structure that covers the goal shape volume.
    \item \textit{Creative Stacked Blocks}: exactly the same as the \textit{Stacked Blocks} task except that the first and last levels of the goal are the only levels shown or "imposed" and the rest of the structure is not explicitly specified, leaving the rest of the goal shape to the imagination of the agent itself; this is considered the most challenging since its it needs the agent to understand how to build stable structures and imagine what can be filled in the middle to connect the two levels in a stable way.
    \item \textit{General}: the goal shape is an arbitrary shape created by initially dropping an arbitrary number of blocks from above the ground and waiting till all blocks come to a rest position where this becomes the goal shape that the agent needs to fill up afterwards.
\end{enumerate}

\begin{table}[!ht]
\centering
\begin{tabular}{p{2.3cm}|p{3cm}|p{3.5cm}|p{3.5cm}}
 Variable & Sub Variable & Space $\mathbf{A}$ & Space $\mathbf{B}$\\
 \hline
 gravity[z] & - & $[-10, -7]$ & $[-7, -4]$\\
 floor friction & - & $[0.3, 0.6]$ & $[0.6, 0.8]$ \\
 stage friction & - & $[0.3, 0.6]$ & $[0.6, 0.8]$ \\
 stage color [rgb] & - & $[0.0, 0.5]^3$ & $[0.5 ,1]^3$ \\
 floor color [rgb] & - & $[0.0, 0.5]^3$ & $[0.5 ,1]^3$ \\
 joint positions & - & $[[-1.57, -1.2, -3.0]^3$, $[-0.69, 0, 0]^3]$ & $[[-0.69, 0, 0]^3$, $[1.0, 1.57, 3.0]^3]$  \\
 block & size &  $[0.055, 0.075]^3$ & $[0.075, 0.095]^3$ \\
 block & color & $[0.0, 0.5]^3$ & $[0.5 ,1]^3$ \\
 block & mass & $[0.015, 0.045]$ & $[0.045 ,0.1]$ \\
 block & position (cylindrical) & $[[0, -\pi, h/2]$, $[0.11, \pi, 0.15]]$ &  $[[0.11, -\pi, h/2]$, $[0.15, \pi, 0.3]]$ \\
 goal cuboid & size & $[0.055, 0.075]^3$ & $[0.075, 0.095]^3$ \\
 goal cuboid & color & $[0.0, 0.5]^3$ & $[0.5 ,1]^3$ \\
 link & color & $[0.0, 0.5]^3$ & $[0.5 ,1]^3$ \\
 link & mass & $[0.015, 0.045]$ & $[0.045 ,0.1]$ \\

\end{tabular}
 \caption{Description of a subset of the high level variables, exposed in CausalWorld, and their corresponding spaces, $h$ refers to the height of the block.}
 \label{tab:causal_world_variables}
\end{table}

\begin{table}[!ht]
\centering
\begin{tabular}{p{2.3cm}|p{3cm}|p{3.5cm}|p{3.5cm}}
 Task Generator & Variable & Space $\mathbf{A}$ & Space $\mathbf{B}$\\
 \hline
 
 Picking & goal height & $[0.08, 0.20]$ & $[0.20 ,0.25]$ \\
 Towers & tower dims & $[[0.08, 0.08, 0.08]$, $[0.12, 0.12, 0.12]]$ &  $[[0.12, 0.12, 0.12]$, $[0.20, 0.20, 0.20]]$ \\

\end{tabular}
 \caption{Example of task generators' specific high level variables, exposed in CausalWorld, and their corresponding spaces. For a full list of each task generators' variables and their corresponding spaces, please refer to the documentation at (https://sites.google.com/view/causal-world/home).}
 \label{tab:causal_world_variables}
\end{table}

\begin{table}[!h]
\begin{tabular}{p{2.5cm}|p{10cm}}
 Task generators & Dense reward\\
 \hline
 Pushing &  $-750\Delta^{t}(o_{1},e) -250\Delta^{t}(o_{1},g_{1}) $\\
 Picking & $-750\Delta^{t}(o_{1},e) - 250\Delta^{t}(o_{1,z},g_{1,z}) - 125\Delta^{t}(o_{1,x,y},g_{1,x,y}) - 0.005||v^{t}-v^{t-1}||$\\
 Pick and Place &  $-750\Delta^{t}(o_{1},e) - 50\Delta^{t}(o_{1,x,y},g_{1,x,y}) - 250(|o^{t}_{1,z} - t| - |o^{t-1}_{1,z} - t|) - 0.005||v^{t}-v^{t-1}||$\\
 Stacking &  $\mathbf{1}_{d^{t}(o_1,e)>0.02} (-750\Delta^{t}(o_{1},e) -250\Delta^{t}(o_{1},g_{1})) + \mathbf{1}_{d^{t}(o_1,e)<0.02} (-750\Delta^{t}(o_{2},e) - 250(|o^{t}_{2,z} - g^{t}_{2,z}| - |o^{t-1}_{1,z} - g^{t}_{2,z}|) - \mathbf{1}_{o^{t}_{2,z} - g^{t}_{2,z} > 0}125\Delta^{t}(o_{2,x,y},g_{2,x,y})) - 0.005||v^{t}-v^{t-1}||$ \\
\end{tabular}
 \caption{Description of the dense rewards applied in our experiments. The following notation was applied: $v^{t} \in \mathbf{R}^3$ joint velocities, $e^{t}_{i} \in \mathbf{R}^3$ i-th end-effector positions, $o^{t}_{i} \in \mathbf{R}^3$ i-th  block position, $g^{t}_{i} \in \mathbf{R}^3$ i-th goal block position, $d^{t}(o,e) = \sum_{i}||e^{t}_i - o^{t}||$ the distance between end-effectors and the  block, $\Delta^{t}_{o,e} = d^{t}(o,e) - d^{t-1}(o,e)$ the distance difference w.r.t. the previous timestep. The target height parameter $t$ for pick and place is 0.15 if block and goal are of different height. Otherwise, $t$ is half the goal height.}
 \label{tab:task-generators}
\end{table}

\clearpage
\section{Training Details}
The experiments were carried out using the \texttt{stable baselines} implementation of PPO, SAC and TD3. We used a 2 layer MLP Policy [256,256] for all the policies. PPO was trained on 20 workers up to 100 million timesteps in parallel and SAC as well as TD3 were trained serially for 10 million timesteps. \\
\\
\begin{table}[!ht]
\begin{tabular}[t]{c|c}
 \multicolumn{2}{c}{PPO} \\
 \hline
 discount & 0.99\\
 batch size & 120000\\
 learning rate & 2.5e-4\\
 entropy coef. & 0.01  \\
 value function coef. & 0.5 \\
 gradient clipping (max) & 0.5 \\
 n minibatches per update & 40 \\
 n training epochs & 4 \\
\end{tabular}
\begin{tabular}[t]{c|c}
 \multicolumn{2}{c}{SAC} \\
 \hline
 discount & 0.95\\
 entropy coeff & 1e-3\\
 batch size & 256\\
 learning rate & 1e-4\\
 target entropy & auto  \\
 buffer size & 1000000 \\
 tau & 0.001 \\
\end{tabular}
\begin{tabular}[t]{c|c}
 \multicolumn{2}{c}{TD3} \\
 \hline
 discount & 0.96\\
 batch size & 128\\
 learning rate & 1e-4\\
 buffer size & 500000 \\
 tau & 0.02 \\

\end{tabular}
  \caption{Learning algorithms hyper parameters used in the baselines experiments.}
 \label{tab:hyper_params}
\end{table}

\begin{figure}[!ht]
\centering
 \includegraphics[width=0.5\linewidth]{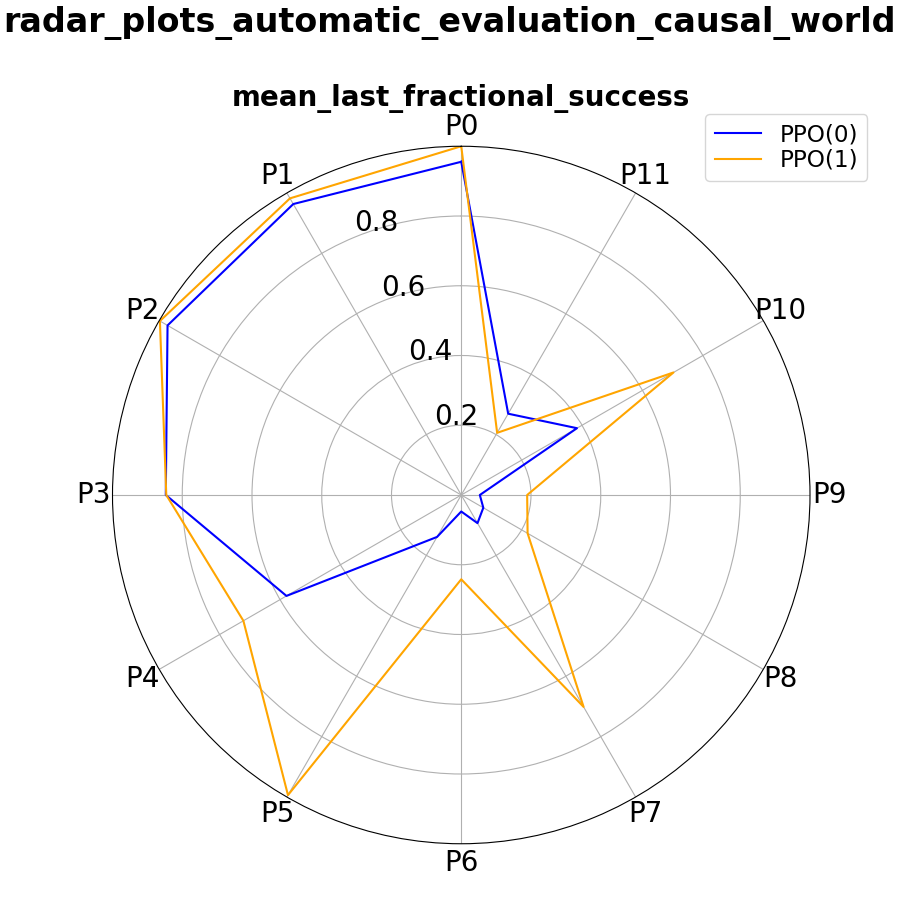}
    \caption{An example of model selection in \texttt{CausalWorld} by evaluating generalization across the various axes using the previously mentioned protocols. Here we compare two agents trained on different curricula using PPO.}
    \label{fig:radar}
\end{figure}

\begin{figure}
    \centering
    \includegraphics[width=\linewidth]{figures/evaluations/pushing_eval_1.pdf}
    \includegraphics[width=\linewidth]{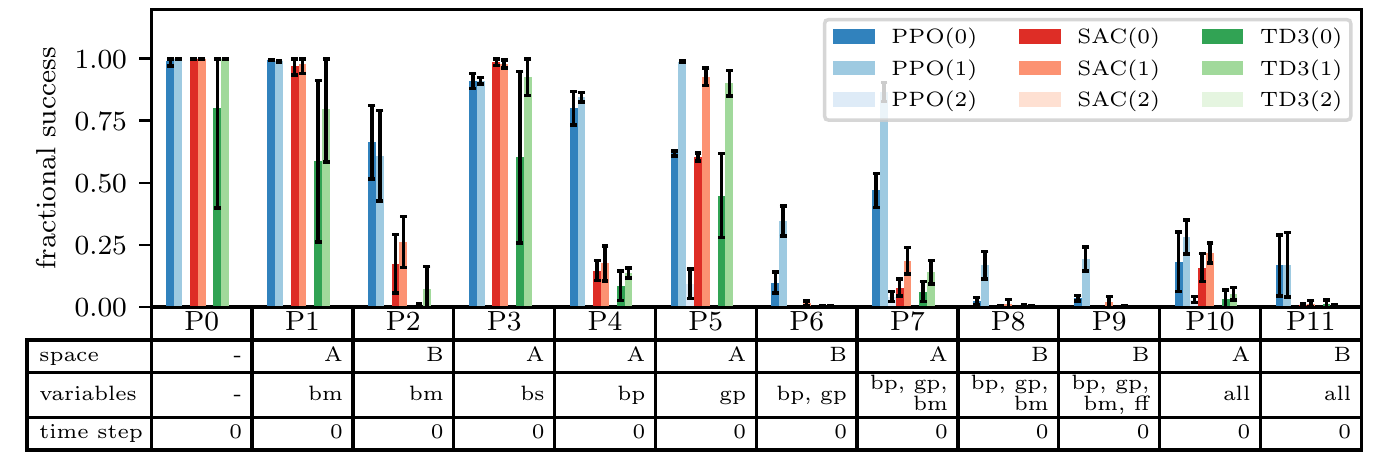}
    \includegraphics[width=\linewidth]{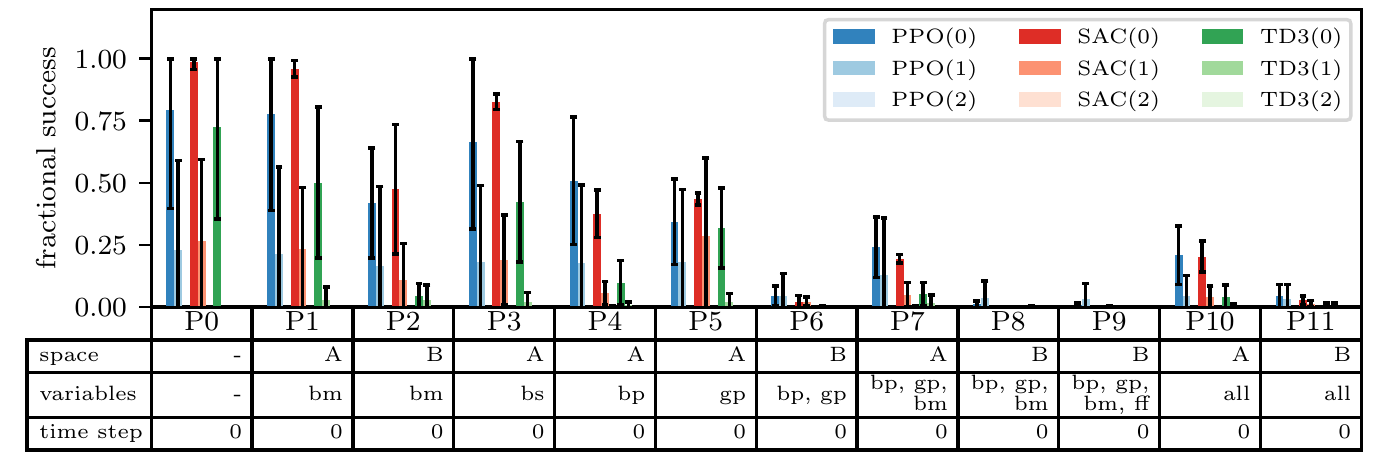}
    \includegraphics[width=\linewidth]{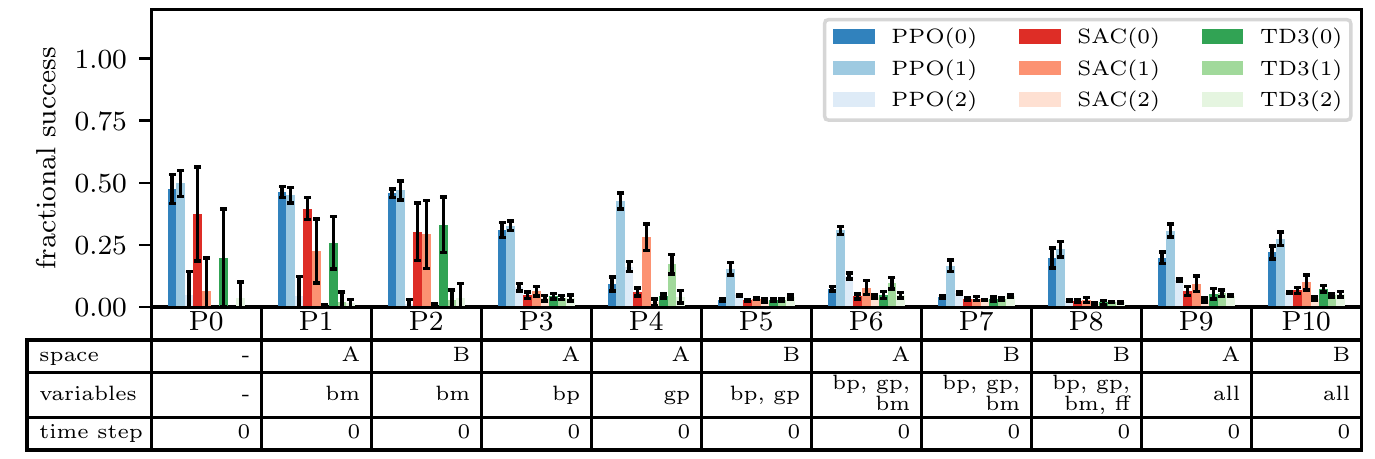}
    \caption{ Evaluation scores, for pushing, picking, pick and place and stacking2 baselines, from top to bottom respectively. Each protocol was evaluated for 200 episodes and each bar is averaged over five models with different random seeds [bp block pose, bm block mass, bs block size, gp goal pose, ff floor friction].}
    \label{fig:appendix_evaluation_protocols_picking}
\end{figure}

\end{document}